\documentclass[letterpaper, 10 pt, conference]{ieeeconf}  

\IEEEoverridecommandlockouts                              

\usepackage{xcolor}
\usepackage{booktabs}
\usepackage{multirow}
\usepackage{makecell}
\usepackage{makecell}
\usepackage{amssymb}
\usepackage{colortbl}
\usepackage{threeparttable}
\usepackage{array}
\usepackage{amsmath}
\usepackage{graphicx}
\usepackage{cite}   
\usepackage[colorlinks,linkcolor=red,anchorcolor=blue,citecolor=blue]{hyperref}
\definecolor{darkgreen}{HTML}{006400}  
\definecolor{darkred}{HTML}{8B0000}    
\newcommand{\greencheck}{$\textcolor{darkgreen}{\checkmark}$}
\newcommand{\redcross}{$\textcolor{darkred}{\times}$}
\definecolor{ccfgray}{RGB}{240,240,240}
\usepackage[T1]{fontenc}
\usepackage{aecompl}

\title{\LARGE \bf
KGCE: Knowledge-Augmented Dual-Graph Evaluator for Cross-Platform Educational Agent Benchmarking with Multimodal Language Models
}

\author{Zixian Liu$^{1}$, Sihao Liu$^{1}$ and Yuqi Zhao$^{1}$*
\thanks{Zixian Liu and Sihao Liu contributed equally to this work.}
\thanks{*This work is supported by the National Natural Science Foundation of China (Nos. 62032016 and 61972292), the Postdoctoral Fellowship Program of China Postdoctoral Science Foundation (No. GZC20240571), and the China Postdoctoral Science Foundation (No. 2024M751052).}
\thanks{$^{1}$Yuqi Zhao is with the Faculty of the School of Computer Science,
        Central China Normal University, Wuhan, China. {\tt\small \{yuqizhao\}@ccnu.edu.cn}}%
}

\begin{document}

\maketitle
\thispagestyle{empty}
\pagestyle{empty}

\begin{abstract}

With the rapid adoption of multimodal large language models (MLMs) in autonomous agents, cross-platform task execution capabilities in educational settings have garnered significant attention. However, existing benchmark frameworks still exhibit notable deficiencies in supporting cross-platform tasks in educational contexts, especially when dealing with school-specific software (such as XiaoYa Intelligent Assistant, HuaShi XiaZi, etc.), where the efficiency of agents often significantly decreases due to a lack of understanding of the structural specifics of these private-domain software. Additionally, current evaluation methods heavily rely on coarse-grained metrics like goal orientation or trajectory matching, making it challenging to capture the detailed execution and efficiency of agents in complex tasks. To address these issues, we propose KGCE (Knowledge-Augmented Dual-Graph Evaluator for Cross-Platform Educational Agent Benchmarking with Multimodal Language Models), a novel benchmarking platform that integrates knowledge base enhancement and a dual-graph evaluation framework. We first constructed a dataset comprising 104 education-related tasks, covering Windows, Android, and cross-platform collaborative tasks. KGCE introduces a dual-graph evaluation framework that decomposes tasks into multiple sub-goals and verifies their completion status, providing fine-grained evaluation metrics. To overcome the execution bottlenecks of existing agents in private-domain tasks, we developed an enhanced agent system incorporating a knowledge base specific to school-specific software. The code can be found at \url{https://github.com/Kinginlife/KGCE}.

\end{abstract}

\section{INTRODUCTION}

The rapid advancement of multimodal large language models (MLMs) is reshaping the capability boundaries of autonomous agents, driving them from single-environment task execution towards cross-platform collaboration \cite{DBLP:journals/corr/abs-2407-01511}. Represented by models like GPT-4o, MLMs integrate visual, linguistic, and action-reasoning capabilities, demonstrating significant potential in general scenarios such as cross-device file transfer and multi-application collaborative operations.

Existing agents have predominantly focused on generic scenarios such as scientific research\cite{DBLP:journals/corr/abs-2310-03302} and code generation\cite{DBLP:conf/acl/ZhangLLSJ24}. However, their performance often declines sharply when transitioning to educational settings due to two major bottlenecks: lack of domain-specific knowledge and misalignment with assessment frameworks. Educational environments pose unique challenges: (1) they heavily rely on school-customized software, characterized by closed private-domain features that lack standardization in interface elements and operational logic; (2) cross-platform tasks require coordination across multiple devices like Windows and Android, involving complex process dependencies and state synchronization; (3) task objectives combine functional requirements with educational significance, demanding agents to exhibit both operational accuracy and cognitive understanding of educational contexts. Current research has yet to effectively address these challenges, thus limiting the practical deployment of educational agents.

\begin{figure}[t]
    \centering
    \includegraphics[width=0.9\linewidth]{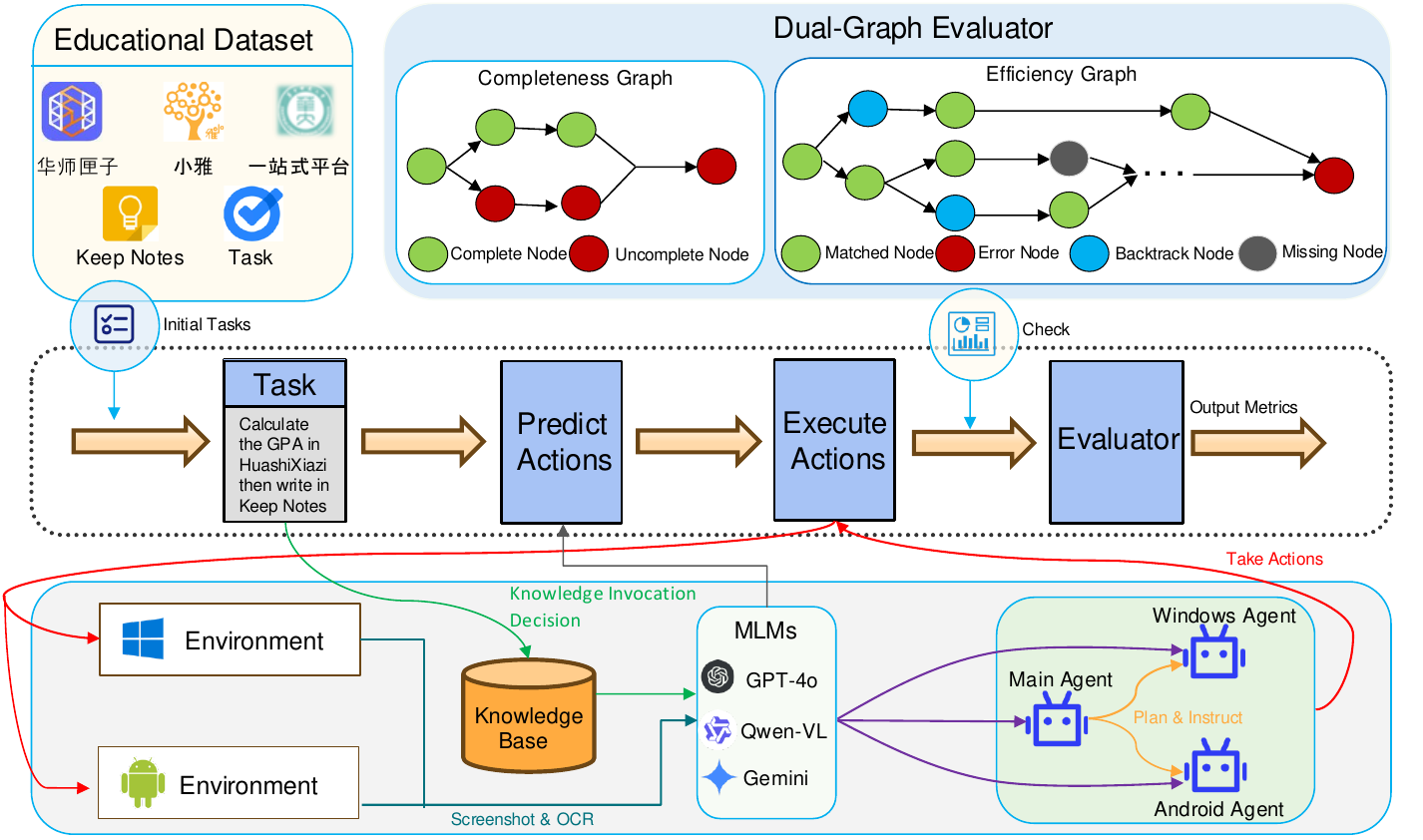}
    \caption{The overall framework of KGCE. The system first generates tasks from educational datasets, then executes them through a pipeline of action prediction, execution, and evaluation. A dual-graph evaluator assesses task completeness and execution efficiency. Based on screenshot and OCR feedback, the system may invoke external knowledge from LLMs (e.g., GPT-4o, Qwen-VL, Gemini) to support cross-environment agents (Windows and Android) in accomplishing complex tasks.}

    \label{fig:enter-label}
\end{figure}

\begin{table*}[t]
        \renewcommand{\arraystretch}{0.9}
                \centering
                \caption{\textbf{ Comparison of existing agent benchmark frameworks.}  
                        }
                \label{tab:environments}
                \footnotesize
                \setlength{\tabcolsep}{3pt}
                \begin{tabular}{@{}lcccccc@{}}
                        \toprule[1.5pt]
                        \textbf{System} & 
                        \makecell{\textbf{Interactive}\\ \textbf{Environment}} & 
                        \textbf{Knowledge} & 
                        \makecell{\textbf{Cross-}\\ \textbf{Platform}} & 
                        \textbf{Evaluation} & 
                        \makecell{\textbf{Task}\\ \textbf{ Construction}} & 
                        \makecell{\textbf{Educational}\\ \textbf{Task}}\\
                        \midrule[1pt]
                        
                        
                        \textsc{MetaGUI\cite{DBLP:conf/emnlp/SunCCDZY22}} & Android & \greencheck & \redcross & Trajectory & Manual & \redcross \\
                        
                        \textsc{AgentBench\cite{DBLP:conf/iclr/0036YZXLL0DMYZ024}} & Multi-isolated & \redcross & \redcross & Multiple & Manual & \redcross \\
                        \textsc{EduAgent\cite{xu2024eduagentgenerativestudentagents}} & Multi-platform & \greencheck & \redcross & Multi-dimensional & LLM+Tools & \greencheck \\
                        \textsc{WebArena\cite{DBLP:conf/iclr/ZhouX0ZLSCOBF0N24}} & Web & \greencheck & \redcross & Goal-based & Template & \redcross \\
                        \textsc{GUICourse\cite{DBLP:journals/corr/abs-2406-11317}} & Desktop/Web GUI & \greencheck & \greencheck & Trajectory & Sub-task Comp & \greencheck \\
                        \textsc{OSWorld\cite{DBLP:conf/nips/XieZCLZCHCSLLXZ24}} & Linux/Windows & \redcross & \redcross & Goal-based & Template & \redcross \\
                        \textsc{AndroidWorld\cite{DBLP:journals/corr/abs-2405-14573}} & Android & \redcross & \redcross & Goal-based & Template & \redcross \\
                        \textsc{EduBenchmark\cite{edubench2025}} & Code/Web & \redcross & \redcross & Multi-dimensional & Template & \greencheck \\
                        \textsc{WORFBench\cite{DBLP:journals/corr/abs-2410-07869}} & Multi & \redcross & \redcross & Graph-based & LLM-inspired & \redcross \\
                        \textsc{CRAB\cite{DBLP:journals/corr/abs-2407-01511}} & Linux\&Android & \redcross & \greencheck & Graph-based & Sub-task Comp & \redcross \\
                        
                        \specialrule{1.2pt}{1pt}{2pt} 
                        \rowcolor{ccfgray}
                        \textbf{\textsc{KGCE}} & Windows\&Android & $\greencheck$ & $\greencheck$ & \textbf{Dual-Graph-based} & Sub-task Comp & \textbf{\greencheck} \\
                        
                        \bottomrule[1.5pt]
                \end{tabular}
                
                \begin{tablenotes}
                    \centering
                        \footnotesize
                        \item[]  \greencheck=Supported, \redcross=Not supported. Cross-platform requires simultaneous multi-device operations.
                \end{tablenotes}
\end{table*}

Existing work exhibits significant limitations in three key areas. Currently, there is a lack of task datasets tailored for educational agents, which hampers research and development in educational scenarios. Existing knowledge graphs \cite{DBLP:journals/csur/ZhongWLPW24} and GUI operation libraries\cite{xu2024eduagentgenerativestudentagents} are primarily designed for general-purpose software and cannot provide structured knowledge support for school-specific systems. Conventional metrics, such as task completion rate and trajectory similarity,  focus solely on macroscopic outcomes. These metrics cannot quantify fine-grained issues like backtracking operations or omissions of critical steps.

To address the the above issues, we propose Knowledge-Augmented Dual-Graph
Evaluator for Cross-Platform Educational Agent Benchmarking with Multimodal Language Models, a cross-platform educational agent benchmarking framework based on knowledge enhancement and dual-graph evaluation. The overall framework is shown in Fig.~\ref{fig:enter-label}.


\textbf{Table \ref{tab:environments}} presents a comparison between KGCE and existing benchmark frameworks. Key features are categorized as: \textit{Interactive Environment} (system's operating context: Web/GUI/Code); \textit{Knowledge} (structured knowledge management with knowledge graphs or dynamic reasoning supported by \greencheck); \textit{Cross-platform} (concurrent multi-device operations requiring OS/device interoperability); 
\textit{Evaluation} (Goal-based: final state verification; Trajectory: action sequence alignment; Graph-based: DAG checkpoint validation); \textit{Task Construction} (Template: predefined patterns; Manual: human-crafted; Sub-task Comp: modular composition);\textit{Educational Task} (curriculum integration or pedagogical assessment via \greencheck).

The main contributions are as follows:
\begin{itemize}
\item We constructed a dataset of 104 educational tasks spanning Windows, Android, and cross-platform collaboration. These tasks involve operations with private-domain software and multi-device coordination workflows, and its dependencies are modeled using a  DAG.

\item For private-domain software, we developed a structured JSON knowledge base. Knowledge from this base is dynamically retrieved and injected into model prompts, significantly improving execution efficiency and success rates in private-domain tasks.
\item We propose a dual-graph evaluation framework consisting of a \textit{Completeness Graph} and an \textit{Efficiency Graph}. This framework introduces eight fine-grained metrics to assess task performance in detail.
\item We validate the effectiveness of the knowledge base across several models, including Qwen-VL-Max-Latest, GPT-4o, and Gemini-2.0-Flash, revealing differences in their dependence on domain-specific knowledge.
\end{itemize}

\section{RELATED WORK}
To comprehensively contextualize our research, we structure the related work into four key dimensions: (1) cross-platform agents driven by large models, (2) task modeling and agents in educational scenarios, (3) knowledge-enhanced agent architectures, (4) agent evaluation methodologies. These dimensions were selected because they collectively address the challenges of building intelligent agents in educational environments. By analyzing these aspects, we aim to highlight the gaps in existing research and position our contributions accordingly.
\subsection{Large Model-Driven Cross-Platform Agents}

In recent years, MLMs have demonstrated significant potential in the field of cross-platform agents. CRAB introduced the first benchmark framework supporting cross-environment tasks, enabling efficient construction and evaluation of complex tasks through subtask composition and a graph-based evaluator. AndroidWorld and OSWorld have established dynamic Android environments and open computer environments, respectively, providing diverse platform support for agent evaluation. However, these cross-platform agent studies do not address the specific support required for education-related private-domain software. Notably, the directed acyclic graph-based task decomposition method proposed by CRAB offers valuable inspiration for our task modeling. Nonetheless, its coarse-grained trajectory-matching evaluation approach falls short of capturing the nuanced execution differences critical to educational scenarios.
\subsection{Task Modeling and Agents in Educational Scenarios}

Research on educational agents faces the dual challenges of complex environments and strong dependence on domain-specific knowledge. EduAgent boosts task efficiency via multimodal interaction, yet only targets general tools. GUICourse\cite{DBLP:journals/corr/abs-2406-11317} improves GUI grounding but ignores cross-platform state sync. EduBench\cite{edubench2025} supplies real-world benchmarks, but its task volume is small and lacks structured knowledge. Current studies exhibit several key shortcomings: (1) Task datasets largely rely on general educational platforms and lack support for school-customized systems; (2) Evaluation metrics focus on macro-level metrics such as task completion rate, which are insufficient for quantifying the optimization level of execution paths. These limitations highlight the critical need for a dedicated evaluation framework tailored to the specific demands of educational scenarios.
\subsection{Knowledge-Augmented Agent Architectures}

Knowledge base augmentation has emerged as an effective paradigm for enhancing agent adaptability across domains. GAT \cite {DBLP:conf/iclr/VelickovicCCRLB18} proposes an attention mechanism to dynamically calculate the importance of nodes in the graph, which is applicable to the priority sorting of nodes in the knowledge atlas and provides a theoretical basis for the priority call mechanism of this research knowledge base. PLaG \cite{DBLP:conf/icml/LinMHY0P24} employed graph structures to enhance task planning capabilities; however, its static knowledge representations are ill-suited for the dynamically evolving nature of educational software. Notably, existing knowledge enhancement methods predominantly rely on general-purpose knowledge graphs\cite{DBLP:conf/acl/SunVSTY20}, lacking structured modeling tailored to private-domain educational software. This essential yet underexplored factor currently constrains the performance of educational agents.
\subsection{Agent Evaluation Methods}

The development of agent evaluation systems is trending from outcome-focused metrics toward process-oriented analysis. AgentBench\cite{DBLP:conf/iclr/0036YZXLL0DMYZ024} established a multidimensional evaluation benchmark, but its API-based validation approach is poorly suited for GUI operation scenarios. DyVal\cite{DBLP:conf/iclr/ZhuC0GY024} introduced a dynamic evaluation framework using a DAG structure to capture task execution increments; however, its discrete state labeling fails to quantify continuous metrics such as the backtracking rate. Regarding fine-grained evaluation, although CRAB’s graph-based evaluator can verify subtask completion states, it lacks the capability to analyze execution path efficiency. Recent studies\cite{DBLP:conf/nips/WuSSSWZFCCXL24} have shown that combining structural analysis with process tracing provides a more accurate reflection of an agent’s cognitive capabilities. These findings offer theoretical support for the design of our dual-graph evaluation framework.

\section{METHOD}
This section provides a detailed introduction to the specific implementation of the cross-platform educational agent benchmark, which combines knowledge base enhancement with a double-layer graph evaluation framework.

\subsection{Educational Dataset Construction} 
While benchmarks cover cross-environment tasks, they miss multi-device collaboration in education. We introduce the first education-focused task set, grounded in real activities at Central China Normal University. Spanning proprietary software, it supports Windows, Android, and cross-platform runs. Inspired by CRAB, we scale via decomposition, templates, and composition: each complex task splits into atomic subtasks, linked in a DAG.  For example, a complex task such as \textit{“Use Xiaoya to check the assignments for the Big Data Technology course and add the task in the Tasks app”} can be broken down into the following subtasks: \textit{“Open the Xiaoya app,” “Enter the Big Data Technology course,” “View the assignment tasks,” “Switch to the Tasks app,”} and \textit{“Add the task.”} These subtasks are organized into a \textbf{ DAG} that models the logical dependencies between them. Each node represents a subtask, and each edge denotes a dependency.

To instantiate concrete tasks, we design task templates using natural language instruction patterns. These templates contain input attributes such as \verb|{app_name}|, \verb|{feature}|, and \verb|{action}|, which can be dynamically replaced with specific values. For example, the template “Open \{feature\} in \{app\_name\}, perform \{action\}, and save it” enables efficient generation of diverse task instances by substituting real values. For complex scenarios, we compose multiple subtask templates to create multi-step tasks. For instance, a cross-platform task may require opening the One-Stop Service Platform on Windows, accessing the message center, and then recording the message content in the Keep Notes app on an Android device.

\begin{figure}[t]
    \centering
    \includegraphics[width=0.9\linewidth]{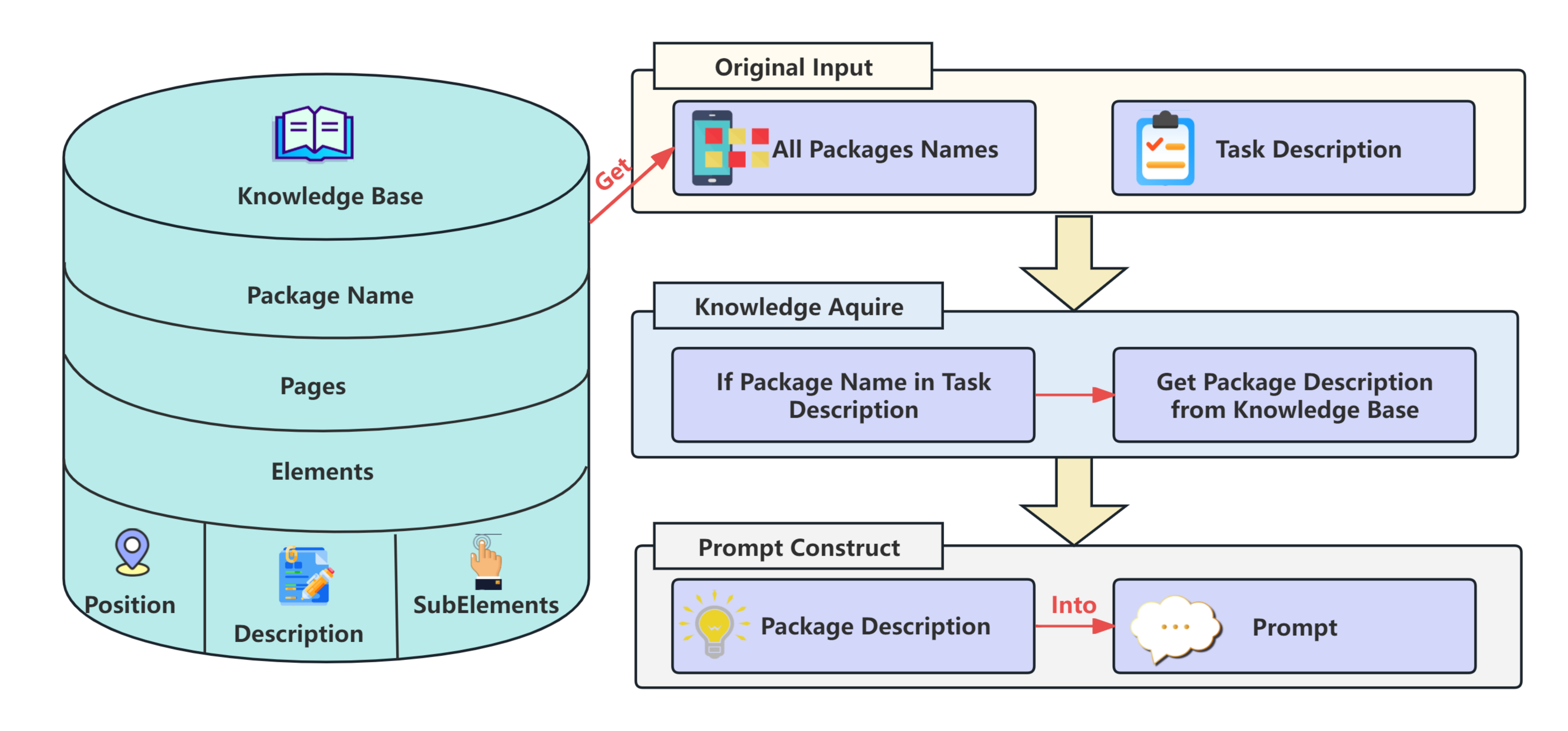}
   \caption{The knowledge base module. Given a task description and package names, the system identifies relevant packages, retrieves their descriptions from the knowledge base, and constructs prompts for the LLM. The knowledge base is organized by packages, pages, and elements, with each element including its position, description, and sub-elements.}

    \label{fig:Knowledge-base}
\end{figure}

To guarantee task feasibility and validity, we rigorously verify each subtask, ensuring it can be executed on the target platform and that its input-output logic is consistent. Ultimately, we construct a dataset of \textbf{104 tasks}, covering applications such as HuaShi XiaZi, Xiaoya Assistant, Keep Notes, and MOOC platforms. This dataset balances task diversity and complexity, laying a solid foundation for subsequent experiments and evaluations.

\subsection{Knowledge Base Construction}

Proprietary software such as Xiaoya Intelligent Assistant presents unique interfaces and workflows that existing MLMs rarely master. To close the gap, we manually interact with each target application to collect software names, page descriptions, UI-element positions, and functional explanations, then serialize the data into a uniform JSON schema. At runtime, a \textbf{Knowledge Invocation Decision} module checks the task description: if the software is mentioned, the corresponding KB records are retrieved and injected into the prompt; otherwise, they are omitted. Fig.~\ref{fig:Knowledge-base} illustrates the full pipeline from manual exploration to structured JSON and prompt augmentation.

\subsection{Dual-Graph Evaluation Framework}
Current coarse-grained, goal-oriented or trajectory-matching metrics inadequately capture the nuanced execution of agents in complex tasks. KGCE proposes a dual-graph evaluation framework—Task Completeness Graph (TCG) and Execution Efficiency Graph (EEG)—to provide fine-grained metrics quantifying both completion quality and execution efficiency.

\subsubsection{Task Completeness Graph}
TCG decomposes a task into sub-goals and builds a dependency graph where each node $v_i$ represents a sub-goal. During execution, we annotate real-time completion status with indicator $I(v_i=1)$. The overall progress is captured by the \textbf{Completion Ratio} $\text{CR} = \sum_{v_i\in V} I(v_i=1)/|V|$. Action-level performance is measured by \textbf{Completeness per Action} $\text{CPA} = \sum_{a_j\in A} I(a_j=1)/|A|$, $I(a_j=1)$ marks whether action $a_j$ is completed.

\subsubsection{Execution Efficiency Graph}
EEG evaluates execution efficiency. \textbf{Backtracking Ratio} $\text{BR} = \text{IO}/\text{ONU}$ quantifies the frequency of backtracking steps (IO) relative to total operations (ONU); lower BR indicates more direct paths. \textbf{Precision} is $\text{CAN}/\text{ANU}$, the ratio of completed actions (CAN) to attempted actions (ANU). \textbf{Recall} measures coverage of essential key steps: $\text{Recall} = \sum_{k_m\in K} I(k_m=1)/|K|$. Precision and Recall are combined into an F1-score. Two exception metrics are recorded: \textbf{Out of Range (OoR)} counts out-of-bounds errors, and \textbf{reach\_max\_step (RMS)} tallies terminations due to reaching the maximum step limit.

\subsection{Dual-Graph Evaluation Framework}
Current evaluation methods predominantly rely on goal-oriented or trajectory-matching coarse-grained metrics, which are insufficient for capturing the execution nuances of agents in complex tasks. KGCE proposes a dual-graph evaluation framework, comprising the Task Completeness Graph and the Execution Efficiency Graph. This framework aims to provide fine-grained evaluation metrics that more accurately quantify the agent’s execution efficiency and task completion quality.

\subsubsection{\textbf{Task Completeness Graph.}}

The Task Completeness Graph evaluates an agent’s ability to achieve individual subgoals within a task.. We decompose each task into multiple subgoals, constructing a dependency graph where each node \{vi\} represents a subgoal. Throughout execution, we monitor and annotate the real-time completion status of each subgoal, using \{I(vi=1)\} to indicate completion (1 for completed, 0 for incomplete). To quantify overall progress and assess whether the agent achieves most subgoals, we define the Completion Ratio (CR):
\begin{equation}
\text{CR} = \frac{\sum_{v_i \in V} I(v_i = 1)}{|V|}.
\label{eq:cr}
\end{equation}

To assess the execution performance at the action level, we propose the \textbf{Completeness per Action (CPA)}:
\begin{equation}
\text{CPA} = \frac{\sum\limits_{a_j \in A} I(a_j = 1)}{|A|}.
\label{eq:cpa}
\end{equation}

Here, ${A}$ is the set of all actions, ${ a_j }$ is the ${ j }$-th action, and ${ I(a_j = 1) }$ is an indicator function that denotes whether action ${ a_j }$ is completed: $I(a_j = 1)$ \text{ indicates whether action } $a_j$ \text{ is completed (1 if completed, 0 otherwise).}

\subsubsection{\textbf{Execution Efficiency Graph.}}

The Execution Efficiency Graph is used to evaluate the efficiency of the agent during task execution. To reflect the agent’s path planning effectiveness, we introduce the \textbf{Backtracking Ratio (BR)}, which measures the frequency of backtracking during execution. A lower BR indicates more direct task completion. It is calculated as IO/ONU, where \textbf{IO} represents the number of backtracking steps due to errors, and \textbf{ ONU} is the total number of operations executed by the agent.
To evaluate the accuracy of action execution, we use the ratio CAN/ANU  to calculate \textbf{Precision}, where \textbf{CAN} denotes the number of completed actions and \textbf{ ANU} is the total number of actions attempted. To assess the coverage of essential steps, we introduce the \textbf{Recall} metric:

\begin{equation}
Recall = \frac{\sum\limits_{k_m \in K} I(k_m = 1)}{|K|},
\label{eq:recall}
\end{equation}
where ${K}$ is the set of all key steps, ${k_m}$ is the ${m}$-th key step, and $I(k_m=1)$ is an indicator function that denotes whether the key step ${k_m}$  is covered: $I(k_m=1)$ indicates whether key step ${k_m}$  is covered (1 if covered, zero otherwise). Combines Precision and Recall to provide a comprehensive evaluation metric F1-score.

Additionally, we introduce two exception metrics: \textbf{Out of Range (OoR)}, which records the number of errors caused by out-of-bounds operations, and \textbf{reach\_max\_step (RMS)}, which records the number of times the agent is terminated due to reaching the maximum step limit.

\section{EXPERIMENT}
In this section, we address the following research questions:
\begin{itemize}
    \item RQ1: Why do we adopt the Dual-Graph Evaluation Framework?
    \item RQ2: What is the impact of incorporating the knowledge module on the final performance?
    \item RQ3: How do different MLMs perform when executing these tasks?
    \item RQ4: For the knowledge base we constructed, which MLM demonstrates the most significant performance improvement? 
    
\end{itemize}

To evaluate the effectiveness of KGCE, we formulated RQ1 to validate the necessity of dual-graph evaluator and fine-grained metrics. RQ2 then uses these metrics to evaluate the impact of knowledge modules on agent performance. On this basis, RQ3 compares the performance of different MLM agents with or without knowledge bases. Finally, RQ4 evaluates the effect of the knowledge base on the performance improvement of each model agent.

\subsection{Compared Methods}
We compare KGCE with two state-of-the-art frameworks: CRAB \cite{DBLP:journals/corr/abs-2407-01511} and WORFBENCH \cite{DBLP:journals/corr/abs-2410-07869}, selected for their relevance to task decomposition and graph-based evaluation, while highlighting gaps addressed by our framework, including comprehensive evaluation, knowledge base utilization, and educational task support.

\textbf{CRAB:} As a benchmark for cross-environment agent evaluation, CRAB introduces a graph-based framework that decomposes tasks into sub-tasks with DAG structures, enabling metrics like CR for multi-platform scenarios. Its interactive GUI environment support and DAG-based evaluator make it a critical baseline for assessing agents across devices, aligning with our focus on cross-platform task execution. However, CRAB lacks a knowledge base, specific support for educational scenarios, and fine-grained metrics, which limits its ability to capture detailed execution efficiency and task completion quality in complex educational tasks.

\textbf{WORFBENCH:} WORFBENCH excels in evaluating complex graph-structured workflows and provides fine-grained metrics for linear and graph planning, which aligns with our interest in task execution analysis. However, it does not address the unique challenges of educational agents, such as domain-specific knowledge requirements and cross-platform task execution in educational software, and lacks macro metrics. Moreover, it only analyzes theoretical results without actual task execution.
\begin{figure}[t]
    \centering
    \includegraphics[width=.85\linewidth]{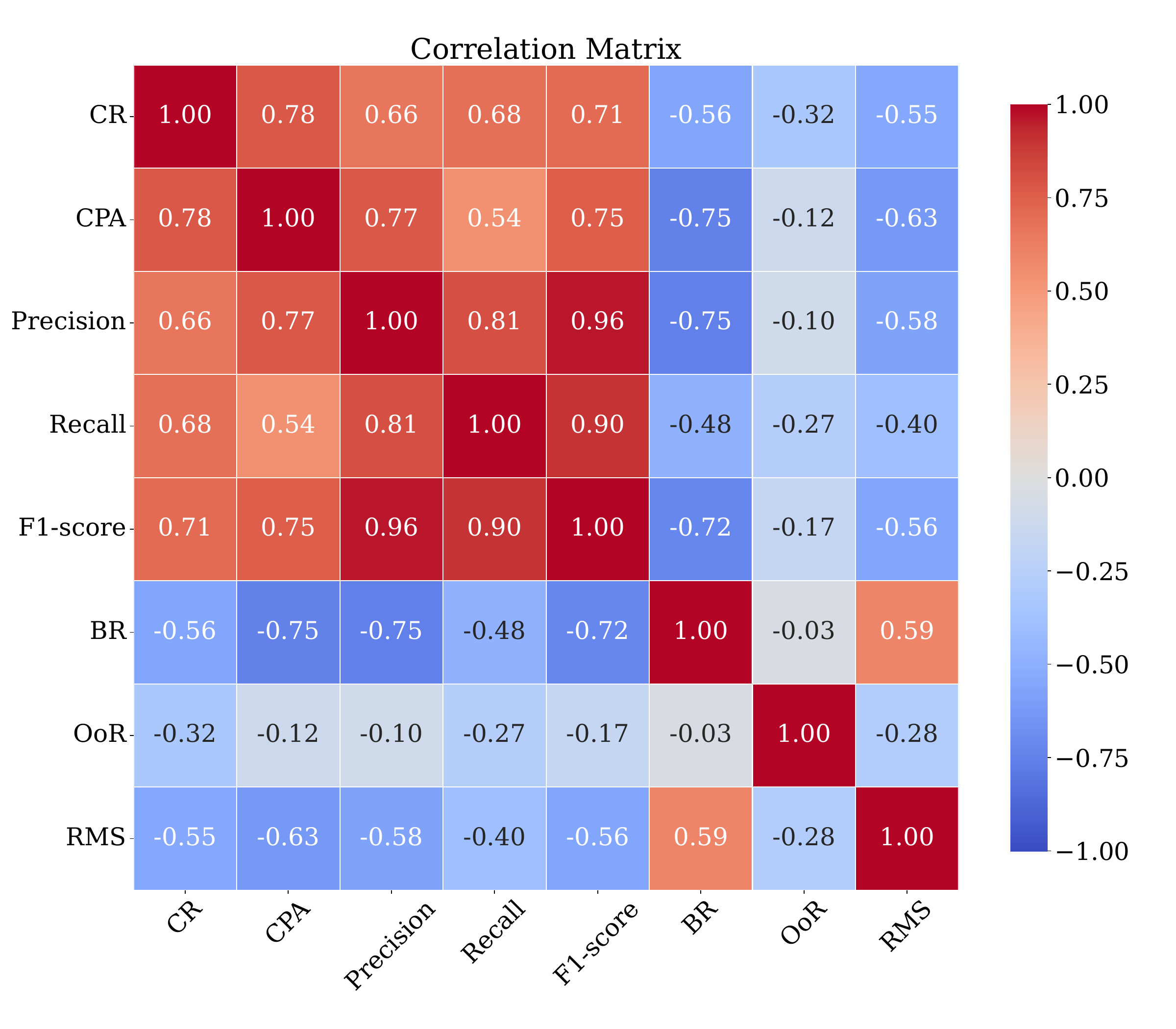}
    \caption{RQ1 Correlation of Metrics.}
    \label{fig: correlation}
\end{figure}
\subsection{RQ1: Validating the Effectiveness of the Dual-Graph Evaluation Framework}
\textbf{Experimental Setup:} To assess the effectiveness of the framework, we conducted 624 task executions across 104 education-related tasks using MLM agents. For each task execution, we recorded all performance metrics generated from the Dual-Graph Evaluation Framework, including CR, CPA, Precision, Recall, F1-score, BR, OoR, and RMS. We then computed Pearson correlation coefficients among all these metrics to evaluate their interdependencies.


\textbf{Experimental Results:}
Fig.~\ref{fig: correlation} presents the correlation-matrix heatmap of all metrics.  Fine-grained indicators from the Dual-Graph Evaluation Framework strongly align with macro-level success: Precision, Recall, and F1-score exhibit high positive correlations with CR, indicating that accurate node selection and comprehensive coverage directly boost task completion.  Conversely, BR, OoR, and RMS are negatively correlated with CR—frequent backtracking (BR), boundary violations (OoR), or exceeding the maximum step limit (RMS) all lower success rates by introducing redundant, erroneous, or incomplete paths.  Together, these micro and macro perspectives provide a clear, actionable diagnosis of agent strengths and weaknesses across multi-path tasks.

\subsection{RQ2: Verify the effectiveness of the knowledge base}
\textbf{Experimental Setup:} To verify the effectiveness of knowledge bases in enhancing the performance of agents, we divided the experiment into two groups: one group is an agent without a knowledge base support, relying solely on the original capabilities of multimodal large models; the other group is an agent with a knowledge base support, which enhances the execution capability of the agent by combining it with the knowledge base. Three models, Qwen-VL-Max-Latest, GPT-4o, and Gemini-2.0-Flash, were used to perform in 104 tasks. In the experiment, we used our dual-graph evaluation framework for performance assessment, including CR, CPA, Precision, Recall, F1-score, BR, OoR, and RMS. The average values of the performance metrics with and without KB were calculated separately.
\begin{table}[t]
        \caption{RQ2 Performance comparison with and without KB}
        \label{tab:performance_comparison}
        \centering
        \begin{tabular}{lcccccc}
                \toprule
                \textbf{Metric} & \textbf{Without KB (\%)} & \textbf{With KB (\%)} & \textbf{Improve (\%)} \\
                \midrule
                CR & 60.02 & 75.26 & +25.39 \\
                CPA & 7.22 & 11.29 & +56.37 \\
                Precision & 24.68 & 32.84 & +33.06 \\
                Recall & 63.87 & 75.79 & +18.66 \\
                F1-score & 33.96 & 44.96 & +32.39 \\
                BR & 52.01 & 41.47 & -20.27 \\
                OoR & 13.42 & 7.54 & -43.81 \\
                RMS & 46.33 & 31.27 & -32.51 \\
                \bottomrule
        \end{tabular}
\end{table}


\begin{figure*}[ht]
    \centering
    \includegraphics[width=0.3\linewidth]{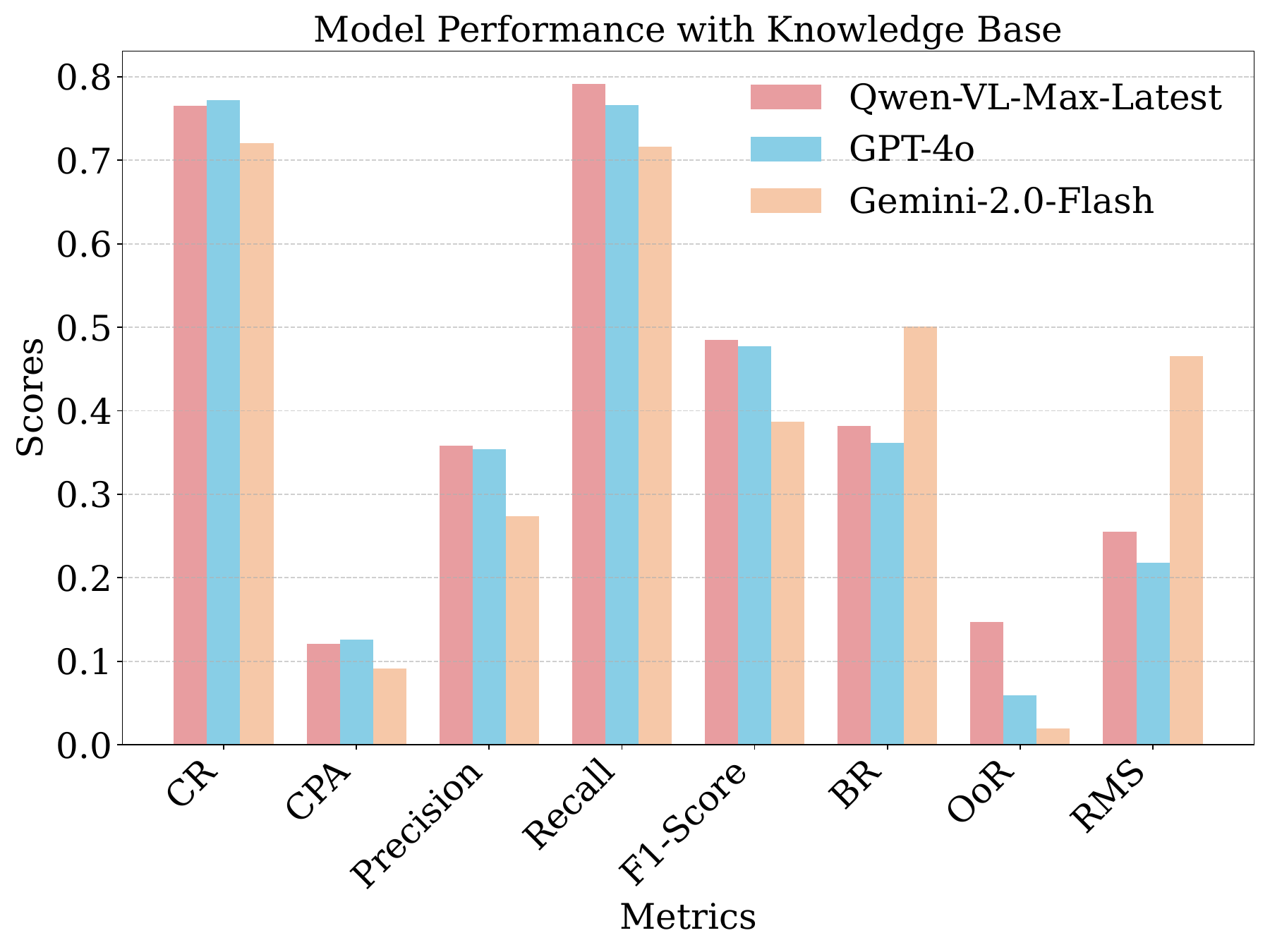}
    \includegraphics[width=0.3\linewidth]{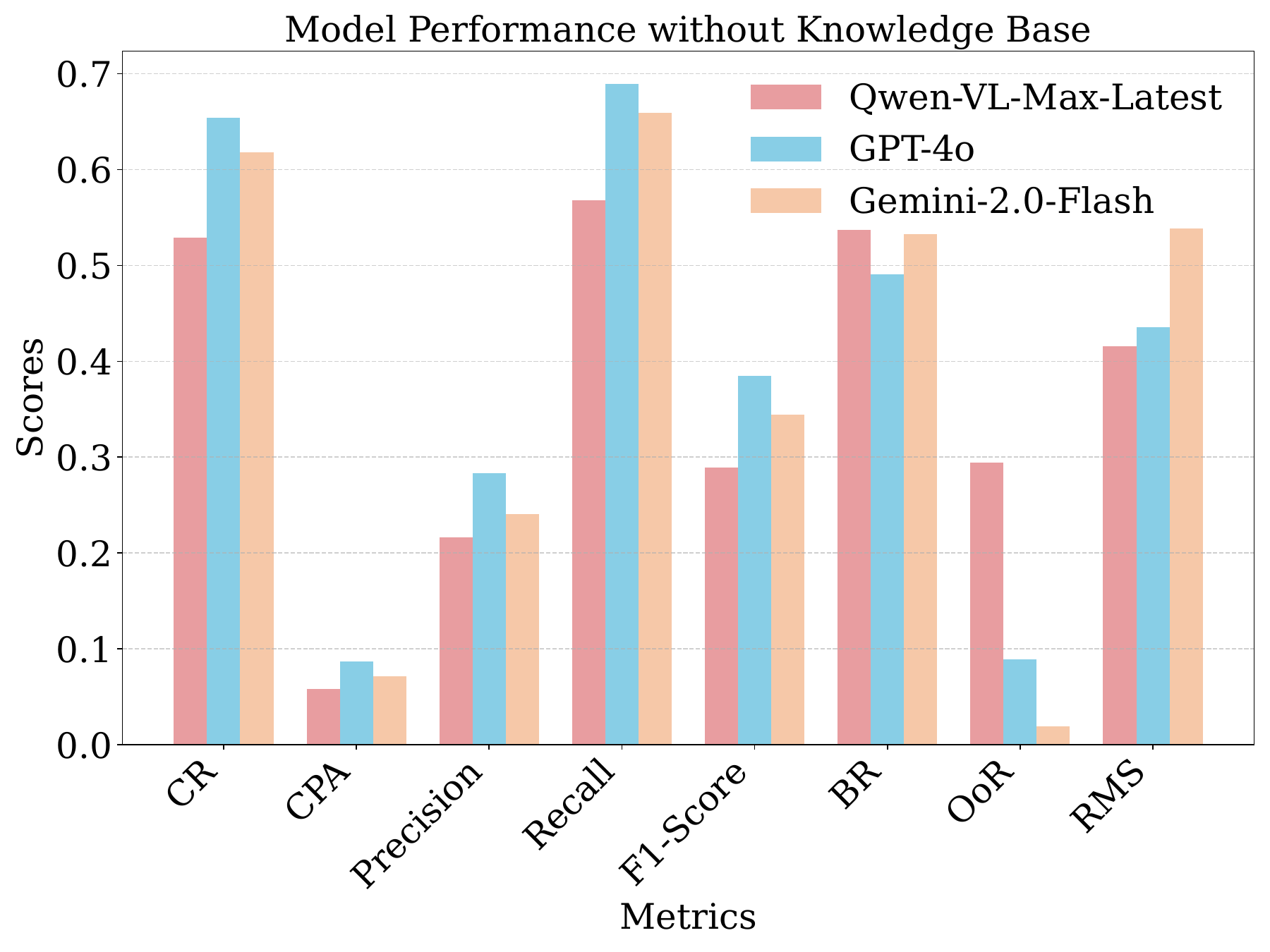}
    \includegraphics[width=0.3\linewidth]{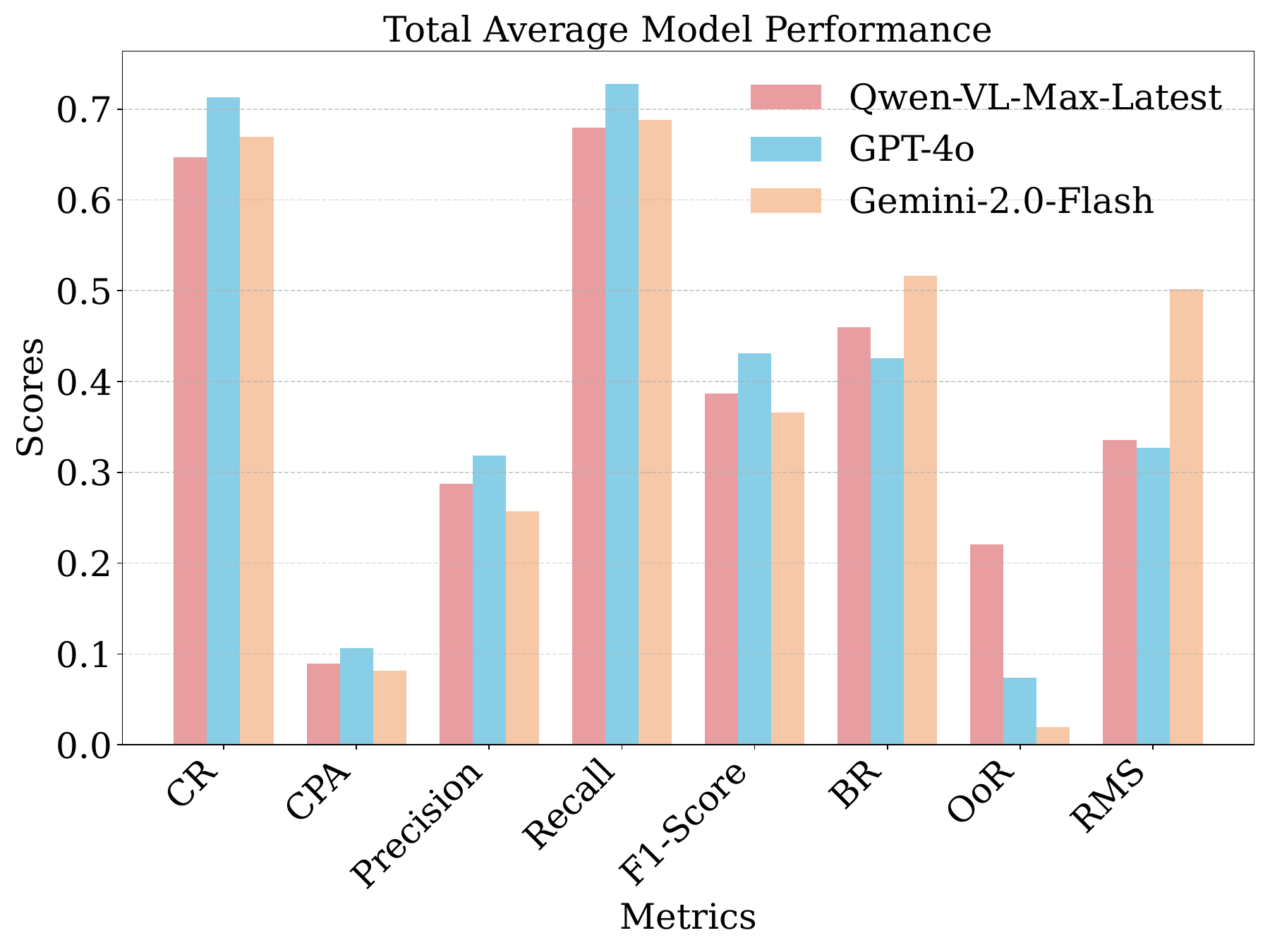}
    \caption{RQ3 Comparison of Multimodal Large Language Models}
    \label{fig:comparison-model-fig}
\end{figure*}

\textbf{Experimental Results:}
Table.~\ref{tab:performance_comparison} summarizes the impact of the knowledge base across all key metrics.
CR rises from 60.02\% to 75.26\% (+25.39\%), CPA from 7.22\% to 11.29\% (+56.37\%); Precision improves from 24.68\% to 32.84\% (+33.06\%), Recall from 63.87\% to 75.79\% (+18.66\%).
At the same time, errors drop sharply: RMS is cut by 32.51\%, OoR by 43.81\%, and BR by 20.27\%.
These gains stem from the knowledge base supplying prior rules, task structures, and refined action prompts that enable the agent to plan shorter, more accurate paths, avoid redundant or out-of-range operations, and cover essential steps. These results indicate that the knowledge base helps agents plan and execute tasks more effectively by providing structured information.

\subsection{RQ3: Evaluate the performance of different large models}

\textbf{Experimental Setup:} To evaluate the performance of different MLM agents on the 104 educational tasks we designed, we used three commercial MLMs: Qwen-VL-Max-Latest, GPT-4o, and Gemini-2.0-Flash. Each model performed tasks with and without a knowledge base, conducting a total of 6 sets of experiments, recording experimental results, and calculating comprehensive performance.

\textbf{Experimental Results:}
The result is shown in Fig~\ref{fig:comparison-model-fig}. From the results, it can be seen that GPT-4o has the best overall performance in terms of task completion rate and execution efficiency. Moreover, with the support of a knowledge base, GPT-4o performs optimally, achieving a CR of 77.21\% and an F1-score of 47.71\%. This indicates that GPT-4o possesses more powerful multimodal processing capabilities, efficient reasoning abilities, deep knowledge retrieval, and semantic understanding. It is capable of fully utilizing the knowledge base and dialogue context, making better use of prior knowledge when handling tasks, avoiding redundant exploration and incorrect operations. In contrast, Qwen-VL-Max-Latest and Gemini-2.0-Flash are relatively weaker in multimodal processing and reasoning capabilities.

\subsection{RQ4: Evaluate the improvement effect of knowledge bases on different large models}

\textbf{Experimental Setup:} To verify the effect of different knowledge bases on the performance improvement of different large model agents, we based on the above experimental metrics calculated to calculate the improvement effect of each metric for each large model.

\begin{table}[t]
    \caption{RQ4 Improvement Comparison of different models}
    \label{tab:model_comparison}
    \centering
    \begin{tabular}{lcccccc}
        \toprule
        \textbf{Model} & \textbf{Metric} & 
        \makecell{\textbf{Without} \\ \textbf{KB (\%)}} &
        \makecell{\textbf{With} \\ \textbf{KB (\%)}} & 
        \makecell{\textbf{Improve} \\ \textbf{(\%)}} \\
        \midrule
        
        \multirow{7}{*}{\parbox{1.3cm}{Qwen-VL-\\Max-Latest}}
        & CR & 52.88 & 76.53 & \textbf{+44.72} \\
        & CPA & 5.82 & 12.09 & \textbf{+107.73} \\
        & Precision & 21.63 & 35.79 & \textbf{+65.46} \\
        & Recall & 56.79 & 79.12 & \textbf{+39.32} \\
        & F1-score & 28.95 & 48.45 & \textbf{+67.43} \\
        & BR & 53.71 & 38.20 & \textbf{-28.88} \\
        & OoR & 29.41 & 14.71 & \textbf{-49.19} \\
        & RMS & 41.58 & 25.49 & -38.70 \\
        
        \midrule
        
        \multirow{7}{*}{GPT-4o} 
        & CR & 65.39 & 77.21 & +18.08 \\
        & CPA & 8.71 & 12.63 & +45.01 \\
        & Precision & 28.33 & 35.37 & +24.85 \\
        & Recall & 68.92 & 76.59 & +11.13 \\
        & F1-score & 38.51 & 47.71 & +23.89 \\
        & BR & 49.08 & 36.12 & -26.41 \\
        & OoR & 8.91 & 5.94 & -33.33 \\
        & RMS & 43.56 & 21.78 & \textbf{-50.00} \\
        
        \midrule
        \multirow{7}{*}{\parbox{1.3cm}{Gemini-\\2.0-Flash}}
        & CR & 61.80 & 72.03 & +16.55 \\
        & CPA & 7.14 & 9.16 & +28.29 \\
        & Precision & 24.08 & 27.35 & +13.58 \\
        & Recall & 65.91 & 71.66 & +8.72 \\
        & F1-score & 34.43 & 38.72 & +12.46 \\
        & BR & 53.25 & 50.08 & -6.07 \\
        & OoR & 1.92 & 1.98 & \textbf{+3.13} \\
        & RMS & 53.85 & 46.53 & -13.59 \\
        
        \bottomrule
    \end{tabular}
\end{table}
\textbf{Experimental Results:}
As can be seen from Table ~\ref{tab:model_comparison}, all three models show a certain improvement in performance after introducing the knowledge base, but compared to them, Qwen-VL-Max-Latest and Gemini-2.0-Flash do not perform as well as GPT-4o overall. Among them, Qwen-VL-Max-Latest shows the most significant improvement after introducing the knowledge base, with the CR increasing from 52.88\% to 76.53\%, while Gemini-2.0-Flash's CR increases from 61.80\% to 72.03\%. This indicates that the knowledge base has a differential impact on the performance improvement of different models, depending on the model's characteristics and requirements. However, Gemini-2.0-Flash's OoR increases by 3.13\% after the introduction of the knowledge base, which theoretically should decrease. This may be due to the conflict between the static rules of the knowledge base and Gemini's dynamic reasoning.

\section{CONCLUSIONS}

This paper proposes KGCE, a cross-platform benchmark framework that integrates knowledge base augmentation and graph-based evaluation to enhance educational agents powered by MLMs. We constructed 104 tasks across Android, Windows, and hybrid platforms, along with a domain-specific software knowledge base. Experiments on Qwen-VL-Max-Latest, GPT-4o, and Gemini-2.0-Flash show consistent performance improvements with knowledge support, with GPT-4o achieving the best results. These findings underscore both the general effectiveness of knowledge augmentation and the impact of model architecture on task adaptability. However, limitations remain in scalability and error detection. Future work will expand task diversity, enrich software coverage, and improve agents' ability to detect and respond to unachievable tasks.

\bibliographystyle{IEEEtran}

\bibliography{ref}

\end{document}